\documentclass[journal]{IEEEtran}

\usepackage{etex}

\usepackage{graphicx} 
\graphicspath{{figs/}}
\usepackage{amsmath} 
\usepackage{amssymb}  
\usepackage{mathtools}

\usepackage{multicol}
\usepackage{placeins}
\usepackage{diagbox}
\usepackage[disable]{todonotes}
\usepackage{adjustbox}
\usepackage{verbatim}
\usepackage{cite}
\usepackage{wrapfig}
\usepackage{booktabs}
\usepackage{url}

\usepackage[noline,linesnumbered]{algorithm2e}
\SetArgSty{textnormal} 
\usepackage{parskip}

\newcommand{\argmax}{\operatornamewithlimits{argmax}}

\begin{document}

\title{Technical Report: The Policy Graph Improvement Algorithm}

\author{Joni Pajarinen
\thanks{J.~Pajarinen is with Aalto University, Finland and Intelligent Autonomous Systems lab, TU Darmstadt, Germany
{\tt\small joni.pajarinen@aalto.fi}}%
}

\maketitle

\begin{abstract}
  Optimizing a partially observable Markov decision process (POMDP)
  policy is challenging. The policy graph improvement (PGI) algorithm
  for POMDPs represents the policy as a fixed size policy graph and
  improves the policy monotonically. Due to the fixed policy size,
  computation time for each improvement iteration is known in
  advance. Moreover, the method allows for compact understandable
  policies. This report describes the technical details of the
  PGI~\cite{pajarinen11} and particle based PGI~\cite{pajarinen17}
  algorithms for POMDPs in a more accessible way than 
  \cite{pajarinen11} or \cite{pajarinen17} allowing practitioners
  and students to understand and implement the algorithms.
\end{abstract}

\IEEEpeerreviewmaketitle

\section{POMDP}

In a POMDP, the agent operates in a world defined by the current state
$s$. However, the agent does not observe $s$ directly but makes
indirect observations about the state of the world. At each time step
$t$, the agent executes an action $a$, receives a reward $R(s,a)$, and
the world transitions to a new state $s^{\prime}$ with probability
$P(s^{\prime}|s,a)$. The agent makes then an observation $o$ with
probability $P(o|s^{\prime},a)$. In a POMDP, the agent can make
optimal decisions based on the complete action-observation history or
a probability distribution over states. The goal of the agent is to
choose actions which maximize the expected total reward
$E[\sum_{t=0}^{T-1} R(s(t), a(t))| \pi, b_0]$ over $T$ time steps,
where $\pi$ is the policy and $b_0$ is the belief, the initial
probability distribution over states.

\section{Policy graph improvement (PGI) algorithm}

\begin{figure}
  \begin{center}
    \includegraphics[width=8cm]{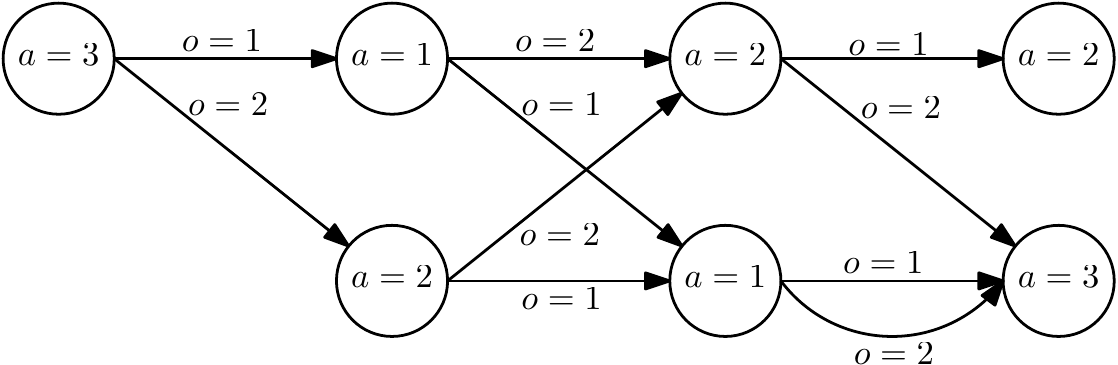}
  \end{center}
  \caption{Example policy graph. The policy graph consists of layers
    of nodes. Each layer corresponds to one time step. Execution
    starts at the far left and proceeds to the right. The agent
    executes actions specified by the policy graph. At time step $t$:
    1) the agent executes action $a_{t,q}$ in policy graph layer $t$,
    where $q$ is the index of the current policy graph node, 2)
    depending on the current world state and action, the agent
    receives a reward and the world transitions to a new state, 3) the
    agent makes an observation $o$ about the new world state, 4) the
    execution moves to the next policy graph node $q^{\prime}$ in the
    next layer $t+1$ along the edge with the matching observation, 5)
    in the next time step $t+1$ the agent executes action
    $a_{t+1,q^{\prime}}$ and so on.}
  \label{fig:policy_graph}
\end{figure}

The PGI algorithm \cite[Algorithm 1]{pajarinen11} improves the value,
that is, the expected total reward over $T$ time steps, of a fixed
size POMDP policy graph (and Dec-POMDP policy graphs)
monotonically. The policy graph is an acyclic graph that consists of
$T$ layers of nodes. A policy graph node executes an action and an
edge, for each possible observation, defines the next node to
transition to. Figure~\ref{fig:policy_graph} shows an example of a
policy graph. See the figure caption for a discussion on how the agent
uses the policy graph for choosing actions.

PGI shares some properties with point based POMDP methods which apply
value iteration with piece wise linear convex (PWLC) value functions
represented as a set of alpha vectors \cite{shani13}. Essentially, PGI
contains an alpha vector at each graph node but it differs from
standard PWLC methods in that it 1) restricts the number of alpha
vectors at each time step, 2) restricts alpha vectors to specific time
steps and allows only alpha vector backups from the next time step, 3)
generates a completely new set of beliefs after changing the
policy. PGI shares the high level idea of alternating between a
forward and backward pass with methods such as differential dynamic
programming (DDP)
\url{https://en.wikipedia.org/wiki/Differential_dynamic_programming}.

\subsection{Notation}

$t$ denotes current time step. $q$ denotes the index of a policy graph
node. $t,q$ denotes graph node $q$ at time step $t$. $A$ is the set of
actions and $a \in A$ is one action. $O$ denotes the set of
observations and $o \in O$ denotes a single observation. $S$ is the
set of states and $s \in S$ is a single state. $s^{\prime}$ denotes
the next time step state and $q^{\prime}$ denotes the next time step
policy graph node.

$P(o, s^{\prime} | s, a) = P(s^{\prime} | s, a) P(o| s^{\prime}, a)$
is the joint transition and observation probability. $b_{t,q}(s)$ is
the \textbf{non-normalized} belief at time step $t$ at policy graph
node $q$. $b$ denotes beliefs at all policy graph nodes.

$\pi$ denotes policy. $\pi$ consists of $P_t(a|q)$ and
$P_t(q^{\prime}|q,o)$. $P_t(a|q)$ denotes the probability to execute
action $a$ at time step $t$ at policy graph node
$q$. $P_t(q^{\prime}|q,o)$ denotes probability to move to policy graph
node $q^{\prime}$ after observing $o$ in policy graph node $q$ at time
step $t$. \emph{Note that in the PGI algorithm $P_t(a|q)$ and
  $P_t(q^{\prime}|q,o)$ are deterministic.} Because of the
deterministic policy we often denote the action at time step $t$ and
at node $q$ with $a_{t,q}$ and the best mapping from action $a$,
observation $o$, current node $q$ to next node as $q_{t+1}(a,o,q)$.

\subsection{PGI for POMDPs}

\begin{figure}
  \centering
  \begin{tabular}{c}
    \framebox[1.1\width]{\includegraphics[width=8.0cm]{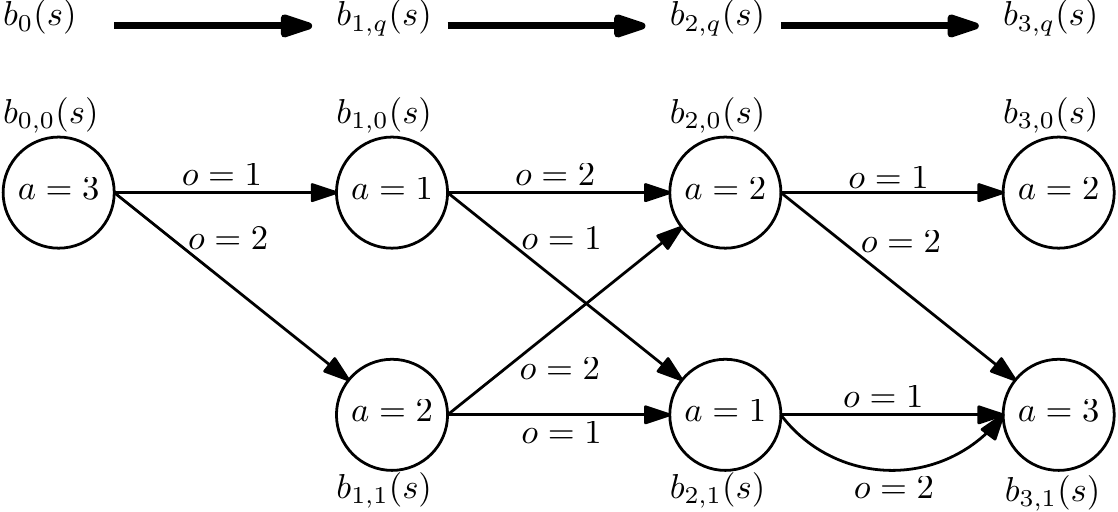}}\\
    \framebox[1.1\width]{\includegraphics[width=8.0cm]{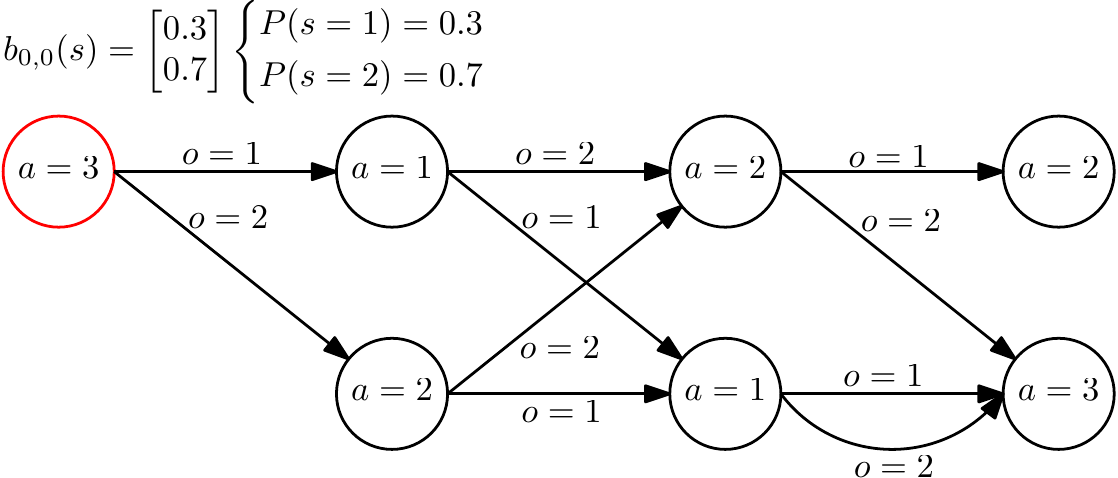}}\\
    \framebox[1.1\width]{\includegraphics[width=8.0cm]{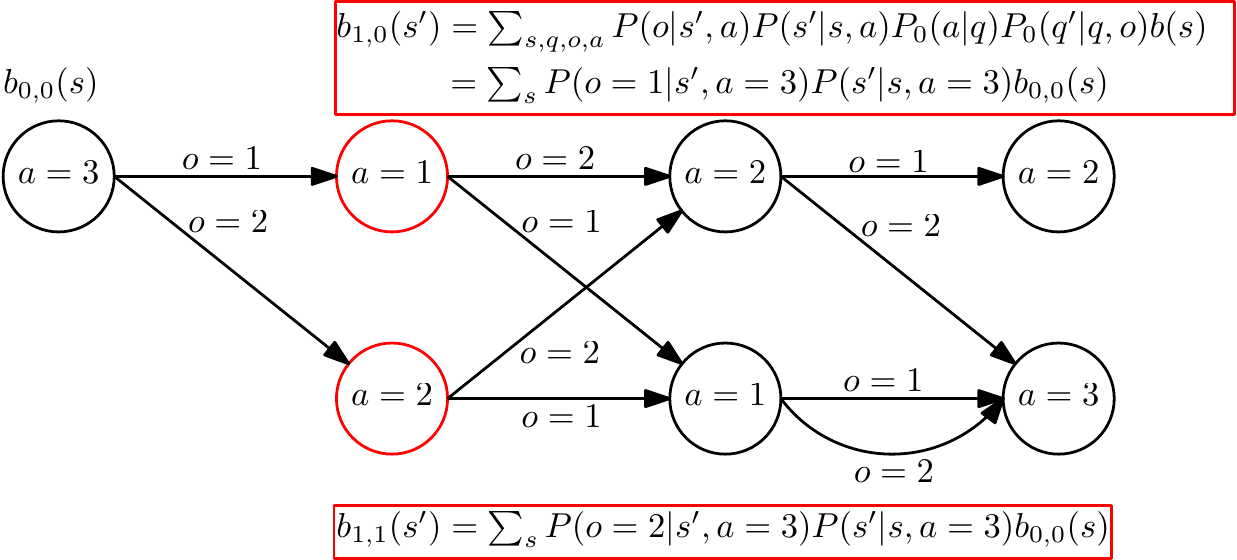}}\\
    \framebox[1.1\width]{\includegraphics[width=8.0cm]{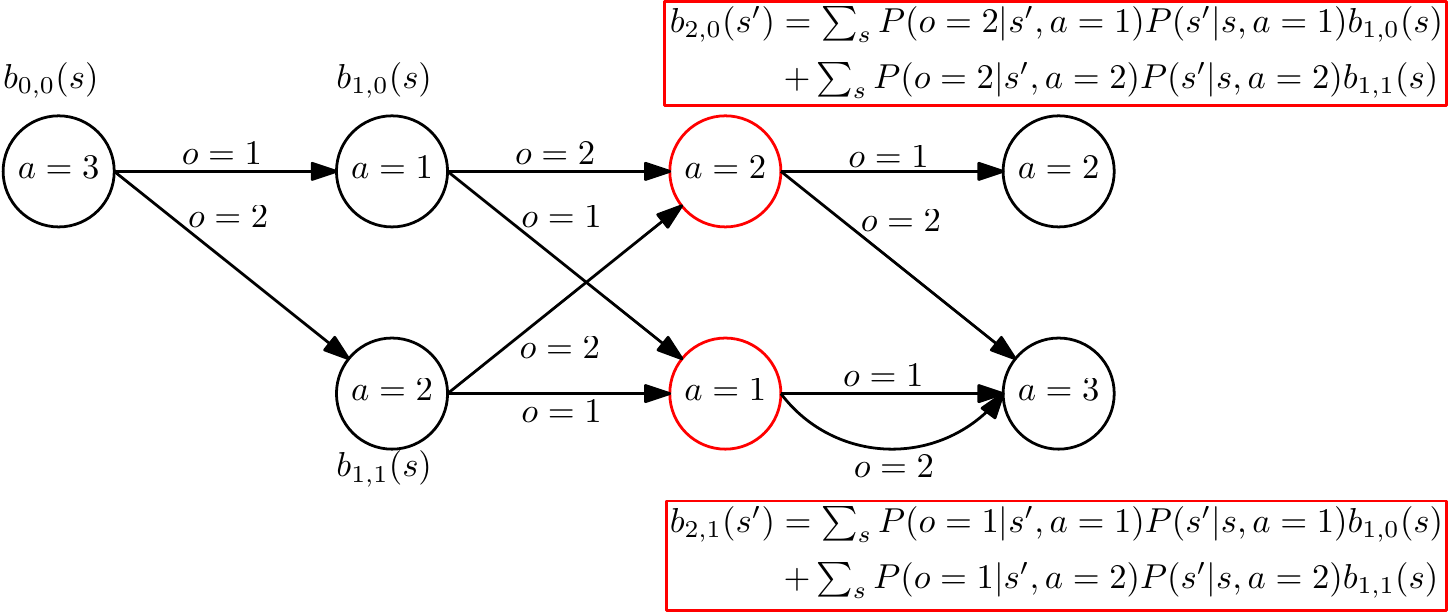}}\\
    \framebox[1.1\width]{\includegraphics[width=8.0cm]{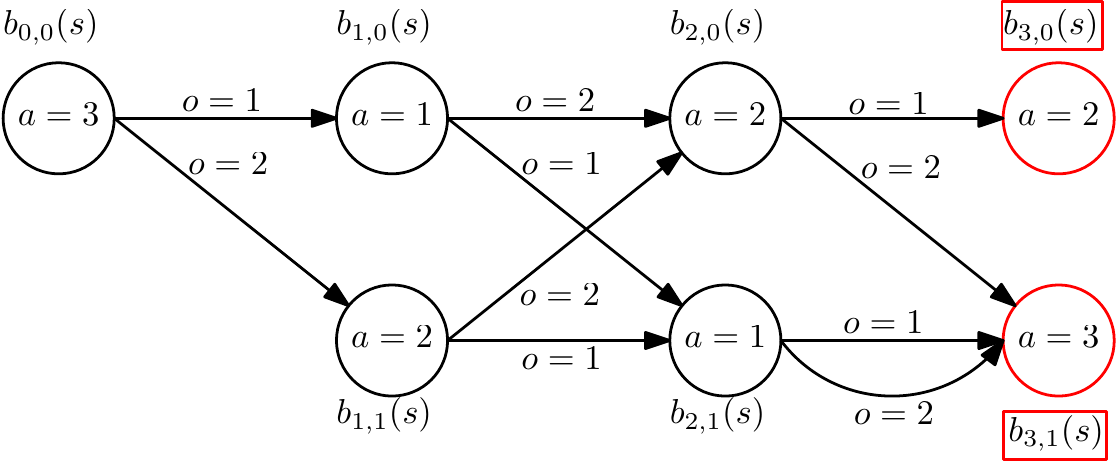}}
  \end{tabular}
  \caption{Illustration of the forward pass procedure. The top figure
    shows a summary of the procedure: project initial belief $b_0(s)$
    through the whole policy graph from left to right using the
    current policy. The following figures show an example of the
    initial belief and the projection steps.}
  \label{fig:forward_pass}
\end{figure}

\begin{figure}
  \centering
  \begin{tabular}{c}
    \framebox[1.1\width]{\includegraphics[width=8.0cm]{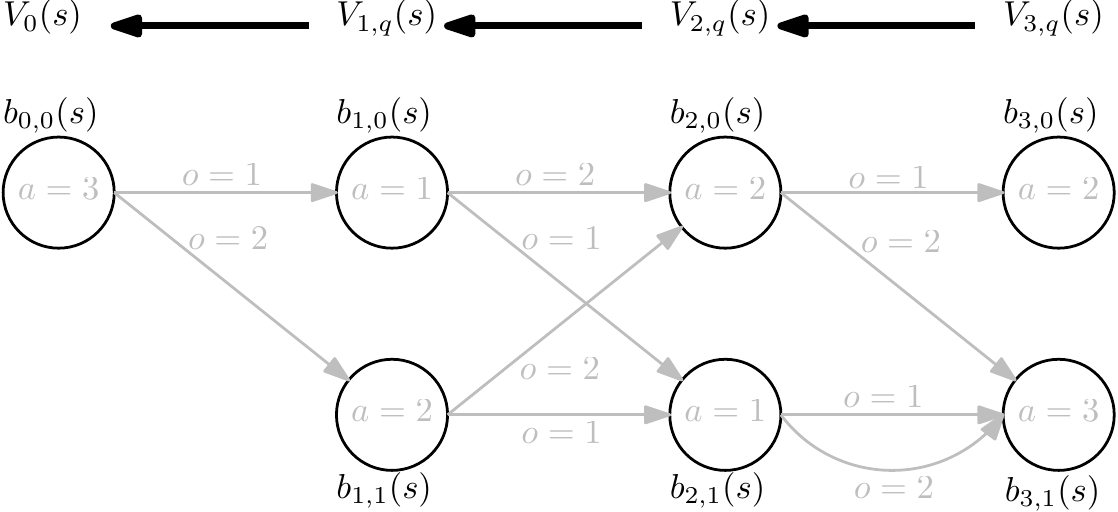}}\\
    \framebox[1.1\width]{\includegraphics[width=8.0cm]{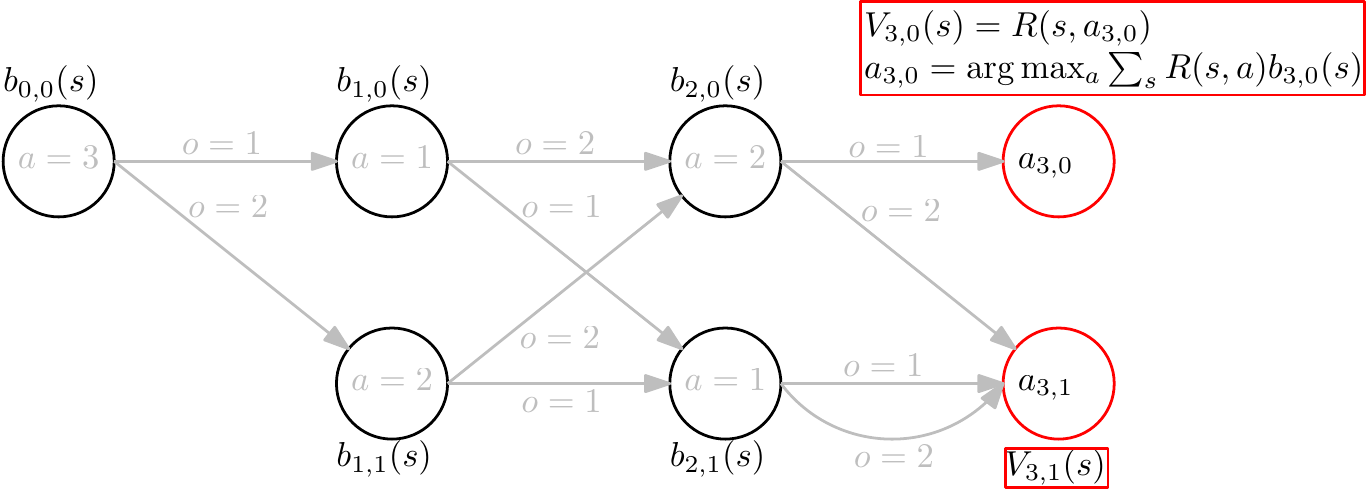}}\\
    \framebox[1.1\width]{\includegraphics[width=8.0cm]{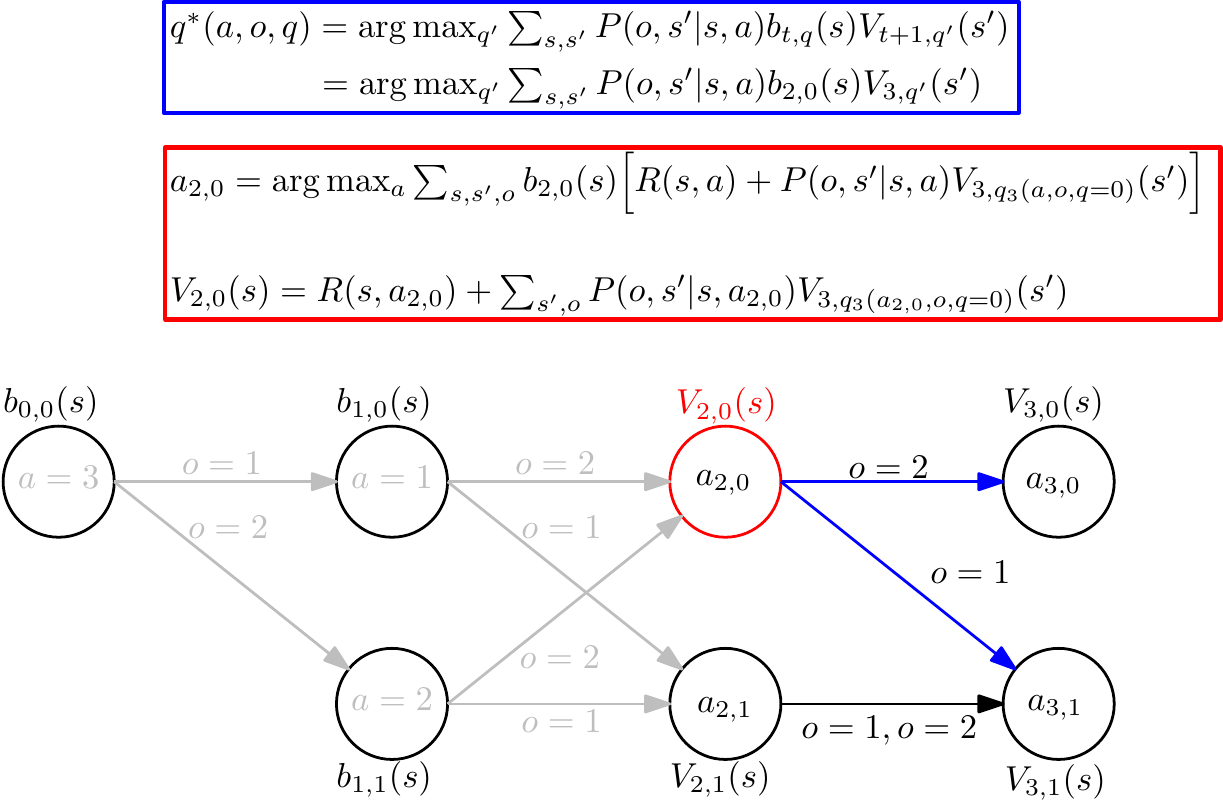}}\\
    \framebox[1.1\width]{\includegraphics[width=8.0cm]{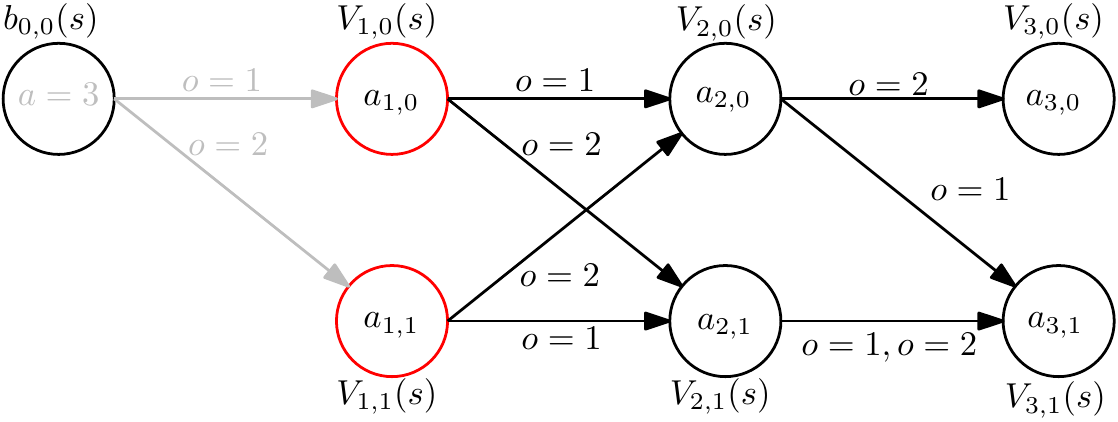}}\\
    \framebox[1.1\width]{\includegraphics[width=8.0cm]{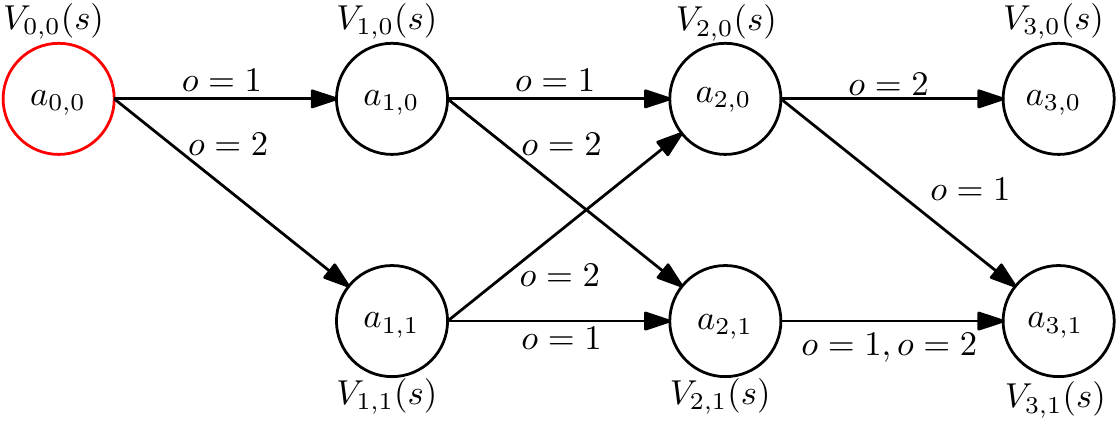}}
  \end{tabular}
  \caption{Illustration of the dynamic programming back pass. The top
    figure shows a summary of the procedure: start from the last
    policy graph layer on the right and update the policy and value
    function at each policy graph layer going from right to left. At
    each node compute best action and for each observation a forward
    edge based on the belief at the node. The following figures show
    an example of the backward pass.}
  \label{fig:back_pass}
\end{figure}

Algorithm~1 defines PGI for POMDPs (without policy
compression~\cite{pajarinen11}). Figure~\ref{fig:forward_pass}
illustrates the forward pass which projects the initial belief, using
the current policy, from the first time step to the last
one. Figure~\ref{fig:back_pass} illustrates the dynamic programming
back pass which optimizes the policy graph for the projected beliefs
starting from the last layer and proceeding to the first one.

\textbf{Policy compression}. Policy compression~\cite{pajarinen11}
recomputes policies for ``redundant'' nodes which get an identical
policy as another policy graph node in the same layer/time step and,
as such, are not useful. A new policy can be computed for the
redundant node using a random belief.

\DontPrintSemicolon
\SetKwProg{Fn}{}{}{}
\begin{algorithm}
  \caption{Monotonic policy graph improvement (PGI) algorithm for POMDPs}
  \label{alg:pgi}
  \SetKwFunction{ForwardPass}{ForwardPass}
  \SetKwFunction{BackPass}{BackPass}
  \SetKwFunction{PGI}{PGI}
  \Fn{$\pi$ = \PGI{$b_0(s)$, $\pi_0$}}{
    \KwIn{Initial belief $b_0(s)$, initial policy $\pi_0 = [P_0(a|q), P_0(q^{\prime}|q,o)]$}
    \KwOut{Optimized policy $\pi$}
    \While{No convergence and time limit not exceeded}{
      $b = $\ForwardPass{$b_0(s)$, $\pi$}\\
      $\pi = $\BackPass{$b$}
    }
  }
\end{algorithm}

\begin{algorithm}
  \caption{PGI forward pass}
  \label{alg:forwardpass}

  \Fn{$b = $\ForwardPass{$b_0(s)$, $\pi$}}{
    $b_{0,0}(s) = b_{0}(s)$\\
    \For{Time step $t=0$ to $T-1$}{
      \ForEach{Policy graph node $q^{\prime}$ at layer $t+1$}{
        $b_{t+1,q^{\prime}}(s^{\prime}) = \sum_{q,o,a,s}
         P(o, s^{\prime} | s, a) P_t(a|q) P_t(q^{\prime}|q,o) b_{t,q}(s)$
      }
    }
  }
\end{algorithm}

\begin{algorithm}
  \caption{PGI dynamic programming back pass}
  \label{alg:backpass}

  \Fn{$\pi = $\BackPass{$b$}}{
    $V_{T+1}(s) = 0$\\
    \For{Time step $t=T$ to $0$}{
      \ForEach{Policy graph node $q$ at layer $t$}{
        // Best next layer graph node $q_{t+1}$\\
        // for $(a,o,q)$:\\
        $q_{t+1}(a,o,q) = \argmax_{q^{\prime}} \sum_{s,s^{\prime}} P(o, s^{\prime} | s, a) b_{t,q}(s) V_{t+1,q^{\prime}}(s^{\prime})$ \\
        // Best action $a_{t,q}^*$ at node $q$:\\
        $a^*_{t,q} = \argmax_a \Big[ \sum_{s} b_{t,q}(s) R(s,a) + \sum_{s,s^{\prime},o} b_{t,q}(s) P(o, s^{\prime} | s, a) V_{t+1,q_{t+1}(a,o,q)}(s^{\prime}) \Big] $ \\
        $P_t(a = a^*_{t,q} | q) = 1$\\
        // Use $a^*_{t,q}$ to get next graph node:\\
        $P_t(q^{\prime} = q_{t+1}(a^*_{t,q},o,q)|q,o) = 1$\\
        // Update value function at node $q$:\\
        {\abovedisplayskip=-1em
          \belowdisplayskip=-0.2em
          \begin{flalign*}
            V_{t,q}(s) =& \sum_a P_t(a|q) \Big[ R(s, a) +
            \sum_{s^{\prime}, o, q^{\prime}}&\\
            & P(s^{\prime},o|s,a) P_t(q^{\prime}|q,o)
            V_{t+1,q^{\prime}}(s^{\prime}) \Big]&
        \end{flalign*}}
      }
    }
  }
\end{algorithm}

\FloatBarrier

\subsection{Particle PGI}

Particle PGI (PPGI)~\cite{pajarinen17} can compute
policies for POMDPs with very large state spaces by using a particle
based approximation of the belief and by approximating the value
function by sampling. Other POMDP methods which use a particle
representation include DESPOT~\cite{somani13}, POMCP~\cite{silver10},
and MCVI~\cite{bai10}. One advantage of PPGI is a fixed size policy
which is incrementally improved instead of growing the policy.

When using a particle based belief, the belief consists of a weighted
set of particles: $b(s) = \sum_{i=1}^N w_i \delta (s, s_i); \sum_i w_i
= 1; 0 \le w_i \le 1$, where $w_i$ is the particle weight and $\delta
(s, s_i)$ is the delta function: $\delta (s, s_i) = 1$ only when $s =
s_i$, otherwise zero. Algorithm~\ref{alg:particle_beliefupdate} shows
how to perform an approximate Bayesian belief update using particles
for a belief given an action and observation.

$N_{t,q}$ denotes the number of particles in belief
$b_{t,q}(s)$. Algorithm~\ref{alg:particle_forwardpass} shows the
particle based forward pass. Algorithm~\ref{alg:particle_backpass}
shows the particle based backwards
pass. Algorithm~\ref{alg:particle_backpass} follows \cite[Algorithm
  1]{bai10}, and, hence, the approximation error bounds in
\cite{bai10} apply.

\begin{algorithm}
  \caption{PPGI forward pass}
  \label{alg:particle_forwardpass}
  \SetKwFunction{ParticleForwardPass}{ParticleForwardPass}
  \SetKwFunction{ParticleBackPass}{ParticleBackPass}

  \Fn{$b = $\ParticleForwardPass{$b_0(s)$, $\pi$}}{
    $b_{0,0}(s) = b_{0}(s)$\\
    \For{Time step $t=0$ to $T-1$}{
      Set $b_{t+1,q^{\prime}}(s^{\prime})$ to an empty set for all $q^{\prime}$\\
      \ForEach{Policy graph node $q$ at layer $t$}{
        \For{$i = 1$ to $N_{t,q}$}{
          // Sample state $s_i$, next state $s_i^{\prime}$,\\
          // and observation $o_i$:\\
          $a = \pi(q)$\\
          $s_i \sim b_{t,q}(s)$\\
          $s_i^{\prime} \sim P(s^{\prime}|s_i,a)$\\
          $o_i \sim P(o|s_i^{\prime},a)$\\
          Add state $s_i^{\prime}$ to $b_{t+1,q_{t+1}(a,o_i,q)}(s^{\prime})$
        }
      }
    }
  }
\end{algorithm}

\begin{algorithm}
  \caption{PPGI dynamic programming back pass}
  \label{alg:particle_backpass}
  \SetKwFunction{Simulate}{Simulate}

  \Fn{$\pi = $\ParticleBackPass{$b$}}{
    \For{Time step $t=T$ to $0$}{
      \ForEach{Policy graph node $q$ at layer $t$}{
        \ForEach{Action $a$}{
          $R_a = 0$\\
          $V_{a,o,q^{\prime}} = 0$ for all $o$, $q^{\prime}$\\
          $V_{a,o} = 0 $ for all $o$\\
          \For{$i = 1$ to $N$}{
            // Sample state $s_i$, next state $s_i^{\prime}$,\\
            // and observation $o$:\\
            $s_i \sim b_{t,q}(s)$\\
            $s_i^{\prime} \sim P(s^{\prime}|s_i,a)$\\
            $o_i \sim P(o|s_i^{\prime},a)$\\
            // Update immediate reward\\
            $R_a = R_a + R(s_i,a)$\\
            \ForEach{Next node $q^{\prime}$}{
              // Simulate future value of $s_i^{\prime}$\\
              $V_{a,o_i,q^{\prime}} = V_{a,o_i,q^{\prime}} +$
              \Simulate{$s_i^{\prime}$,$t+1$,$q^{\prime}$,$\pi$}\\
            }
          }
          \ForEach{Observation $o$}{
            // Next layer node $q_{t+1}(a,o,q)$ for\\
            // current node $q$ and each\\
            // $a$, $o$ combination:\\
            $q_{t+1}(a,o,q) = \argmax_{q^{\prime}} V_{a,o,q^{\prime}}$\\
            $V_{a,o} = V_{a,o,q_{t+1}(a,o,q)}$\\
          }
          $V_a = (R_a + \sum_o V_{a,o}) / N$\\
        }
        // Optimized action for node at $t,q$:\\
        $a_{t,q} = \argmax_a V_a$\\
      }
    }
  }
\end{algorithm}

\begin{algorithm}
  \caption{Simulate future state value given policy}
  \label{alg:simulate}

  \Fn{$V = $\Simulate{$s_i$, $t$, $q$, $\pi$}}{
    $V = R(s_i,a_{t,q})$\\
    \For{Time step $t$ to $T-1$}{
      $s_i^{\prime} \sim P(s^{\prime}|s_i,a_{t,q})$\\
      $o_i \sim P(o|s_i^{\prime},a_{t,q})$\\
      $q^{\prime} = q_{t+1}(a_{t,q},o_i,q)$\\
      $V = V + R(s^{\prime},a_{t+1,q^{\prime}})$\\
      // Move policy node and state to next time step:\\
      $q = q^{\prime}$, $s_i = s_i^{\prime}$\\
    }
  }
\end{algorithm}

\begin{algorithm}
  \caption{Particle belief update}
  \label{alg:particle_beliefupdate}
  \SetKwFunction{ParticleBeliefUpdate}{ParticleBeliefUpdate}
  \Fn{$b(s^{\prime}) = $\ParticleBeliefUpdate{$b(s)$, $a$, $o$}}{
    Set $b(s^{\prime})$ to an empty set\\
    \For{$i = 1$ to $|b(s)|$}{
      $w_i, s_i = b_i(s)$\\
      $s_i^{\prime} \sim P(s^{\prime}|s_i,a)$\\
      $w_i^{\prime} = w_i P(o|s_i^{\prime},a)$\\
      Add $w_i^{\prime}, s_i^{\prime}$ to $b(s^{\prime})$\\
    }
    Normalize $b(s^{\prime})$\\
  }
\end{algorithm}

\FloatBarrier

\bibliographystyle{IEEEtran}
\bibliography{root}

\begin{thebibliography}{1}
\providecommand{\url}[1]{#1}
\csname url@rmstyle\endcsname
\providecommand{\newblock}{\relax}
\providecommand{\bibinfo}[2]{#2}
\providecommand\BIBentrySTDinterwordspacing{\spaceskip=0pt\relax}
\providecommand\BIBentryALTinterwordstretchfactor{4}
\providecommand\BIBentryALTinterwordspacing{\spaceskip=\fontdimen2\font plus
\BIBentryALTinterwordstretchfactor\fontdimen3\font minus
  \fontdimen4\font\relax}
\providecommand\BIBforeignlanguage[2]{{%
\expandafter\ifx\csname l@#1\endcsname\relax
\typeout{** WARNING: IEEEtran.bst: No hyphenation pattern has been}%
\typeout{** loaded for the language `#1'. Using the pattern for}%
\typeout{** the default language instead.}%
\else
\language=\csname l@#1\endcsname
\fi
#2}}

\bibitem{pajarinen11}
J.~Pajarinen and J.~Peltonen, ``{Periodic finite state controllers for
  efficient POMDP and DEC-POMDP planning},'' in \emph{Advances in Neural
  Information Processing Systems (NIPS)}, 2011, pp. 2636--2644.

\bibitem{pajarinen17}
J.~Pajarinen and V.~Kyrki, ``{Robotic manipulation of multiple objects as a
  POMDP},'' \emph{Artificial Intelligence}, 2017.

\bibitem{shani13}
G.~Shani, J.~Pineau, and R.~Kaplow, ``{A survey of point-based POMDP
  solvers},'' \emph{Autonomous Agents and Multi-Agent Systems}, vol.~27, no.~1,
  pp. 1--51, 2013.

\bibitem{somani13}
A.~Somani, N.~Ye, D.~Hsu, and W.~S. Lee, ``{DESPOT: Online POMDP planning with
  regularization},'' in \emph{Advances in Neural Information Processing Systems
  (NIPS)}, 2013, pp. 1772--1780.

\bibitem{silver10}
D.~Silver and J.~Veness, ``{Monte-Carlo planning in large POMDPs},'' in
  \emph{Advances in Neural Information Processing Systems (NIPS)}, 2010, pp.
  2164--2172.

\bibitem{bai10}
H.~Bai, D.~Hsu, W.~S. Lee, and V.~A. Ngo, ``{Monte Carlo value iteration for
  continuous-state POMDPs},'' in \emph{Algorithmic foundations of robotics
  IX}.\hskip 1em plus 0.5em minus 0.4em\relax Springer, 2010, pp. 175--191.

\end{thebibliography}

\end{document}